\documentclass[sigconf,9pt]{acmart}
\renewcommand\footnotetextcopyrightpermission[1]{}
\setcopyright{none}

\usepackage{amsmath,subcaption}
\usepackage{booktabs}
\usepackage{bbm,dsfont}
\usepackage{algorithm}
\usepackage{algpseudocode}
\usepackage{makecell}
\usepackage{multicol}
\usepackage{soul}
\usepackage{pifont}
\usepackage{multirow}
\usepackage{graphicx}
\AtBeginDocument{%
  }

\usepackage[colorinlistoftodos,textsize=tiny
			]{todonotes}

\begin{document}

\title{IISE PG\&E Energy Analytics Challenge 2025: Hourly-Binned Regression Models Beat Transformers in Load Forecasting}

 \author{
 {\rm Millend Roy\textsuperscript{*}}, 
 {\rm Vladimir Pyltsov\textsuperscript{*}}, 
 {\rm  Yinbo Hu}\\
 Columbia University, New York, NY, USA
} 

\begin{abstract}
Accurate electricity load forecasting is essential for grid stability, resource optimization, and renewable energy integration. While transformer-based deep learning models like TimeGPT have gained traction in time-series forecasting, their effectiveness in long-term electricity load prediction remains uncertain. This study evaluates forecasting models ranging from classical regression techniques to advanced deep learning architectures using data from the ESD 2025 competition. The dataset includes two years of historical electricity load data, alongside temperature and global horizontal irradiance (GHI) across five sites, with a one-day-ahead forecasting horizon. Since actual test set load values remain undisclosed, leveraging predicted values would accumulate errors, making this a long-term forecasting challenge. We employ (i) Principal Component Analysis (PCA) for dimensionality reduction and (ii) frame the task as a regression problem, using temperature and GHI as covariates to predict load for each hour, (iii) ultimately stacking 24 models to generate yearly forecasts.

Our results reveal that deep learning models, including TimeGPT, fail to consistently outperform simpler statistical and machine learning approaches due to the limited availability of training data and exogenous variables. In contrast, \textbf{XGBoost, with minimal feature engineering, delivers the lowest error rates across all test cases while maintaining computational efficiency.} This highlights the limitations of deep learning in long-term electricity forecasting and reinforces the importance of model selection based on dataset characteristics rather than complexity. Our study provides insights into practical forecasting applications and contributes to the ongoing discussion on the trade-offs between traditional and modern forecasting methods.

\end{abstract}

\maketitle
\pagestyle{plain}
\renewcommand{\thefootnote}{\fnsymbol{footnote}}
\footnotetext[1]{Authors have equal contribution.}
\renewcommand{\thefootnote}{\arabic{footnote}}

\section{Introduction}


Accurate time-series forecasting plays a critical role in the energy sector, impacting grid stability, market settlements \cite{weron2014electricity}, renewable energy integration \cite{kang2020renewable}. Independent System Operators (ISOs) and Load Serving Entities (LSEs)  \cite{nerc2022reliability, ferc2021staff} rely on forecasting models for capacity planning \cite{SAHU2023109025}, congestion management, and ancillary service procurement \cite{weron2014electricity}. With the increasing penetration of renewable energy sources such as wind and solar, forecasting methodologies must accommodate their inherent intermittency and uncertainty \cite{arias2022probabilistic}. Advances in machine learning (ML) \cite{keles2016extended} and deep learning (DL)  \cite{lago2018forecasting} have introduced sophisticated models, leveraging vast datasets from smart meters, weather stations, and economic indicators \cite{haben2016analysis} to improve forecast accuracy.

Despite the growing adoption of deep learning models such as recurrent neural networks (RNNs), transformer-based architectures, and attention mechanisms for time-series forecasting, their effectiveness in long-term electricity load prediction remains uncertain \cite{hyndman2020large}. Studies have shown that while these models excel in short-term predictions, they often deteriorate over extended forecasting horizons due to error accumulation and distribution shifts \cite{lai2018modeling}. Similarly, prior research highlights that complex ML models frequently underperform compared to simpler statistical approaches like ARIMA and exponential smoothing when evaluated on diverse datasets \cite{green2015simple}, primarily due to high sensitivity to hyperparameter tuning, lack of interpretability, and dependence on extensive training data. 

Our study aims to address this gap by evaluating various forecasting methodologies for electricity load prediction under constrained conditions. The \textsc{ESD 2025 competition dataset} presents a unique challenge: only two years of historical electricity load; and temperature and global horizontal irradiance (GHI) from five different sites are provided, with a forecasting horizon of one-day ahead. Moreover, the actual load values for the test dataset remain undisclosed, preventing the use of autoregressive components and making it a long-term forecasting challenge. Traditional deep learning methods often rely on lagged variables for feature extraction, but given the problem constraints, alternative modeling approaches must be explored.

To tackle this challenge, we develop a robust forecasting framework that systematically compares a range of models, from classical regression techniques to state-of-the-art deep learning architectures. Our key contributions include:
\begin{itemize}
    \item Reframing the problem as a regression task by utilizing PCA-transformed exogenous variables (GHI and temperature) to predict hourly load values.
    \item Employing a stacking approach with 24 independent hourly models to construct yearly load forecasts.
    \item Systematic evaluation through multiple validation strategies, including cross-validation and holdout methods, to ensure model robustness.
    \item Exploration of lagged and leading feature engineering techniques for the exogenous variables only to assess their impact on predictive performance.
    \item Benchmarking model performance against deep learning architectures such as TimeGPT, demonstrating that simpler ML models, particularly XGBoost, outperform more complex methods in this constrained setting.
\end{itemize}

\label{introduction}

\section{Related Work}
\textbf{Comparing Deep Learning, Machine Learning and Hybrid approaches:}
0The integration of deep learning (DL) and machine learning (ML) techniques in time-series forecasting has gained significant traction, showing both promising results and inherent limitations. Deep learning approaches such as recurrent neural networks (RNNs), temporal convolutional networks (TCNs), and transformer-based models have demonstrated their ability to capture complex temporal dependencies in electricity load data \cite{Lanka,benidis2020neural,salinas2020deepar}. However, studies like Makridakis \textit{et al.} \cite{makridakis2018statistical} highlight that complex ML models often underperform compared to simpler statistical approaches such as ARIMA and exponential smoothing, particularly when evaluated across diverse datasets. Their work revealed that these models are highly sensitive to hyperparameter tuning, require extensive training data, and can struggle with interpretability. Green \textit{et al.} \cite{green2015simple} found that in 97 comparisons, complex models resulted in higher errors than simpler models in 81\% of cases. Similarly, the M4 Forecasting Competition results analyzed by Gilliland \cite{gilliland2019m4} indicate that hybrid models, combining statistical methods with selective ML components, frequently outperform pure machine learning models.

\textbf{Challenges in Long-Term Forecasting for Electricity Load :}
A major concern in forecasting literature is the performance degradation of advanced models over extended forecasting horizons. Hyndman \textit{et al.} \cite{hyndman2020large} documented how sophisticated neural forecasting models, which perform well for short-term predictions, exhibit exponential error growth as the forecast horizon extends. The M5 competition results \cite{makridakis2022m5} further confirm this, showing the unstable performance of deep learning models across different time horizons. Long-term forecasting is particularly challenging in energy markets, where errors accumulate in autoregressive frameworks, and shifting distributions complicate predictions. Lai \textit{et al.} \cite{lai2018modeling} attribute these issues to error accumulation and regime shifts over time.

\textbf{Advancements in Forecasting: Current State-of-the-Art:}
Electricity load forecasting involves diverse methodologies, each with specific advantages. (1) Probabilistic forecasting methods provide predictive distributions rather than point estimates, making them useful for uncertainty-aware planning \cite{rubaszewski2022probabilistic, nowotarski2013computing}. (2) Traditional regression-based techniques such as ARIMA and exponential smoothing remain widely used due to their interpretability and minimal data requirements \cite{petropoulos2022forecasting}. (3) ML models, including random forests, gradient boosting machines, and support vector regression, offer flexibility in capturing nonlinear dependencies without requiring explicit model assumptions \cite{sezer2020financial}. Recently, (4) deep learning architectures like DeepAR \cite{salinas2020deepar}, N-BEATS \cite{oreshkin2020n}, and transformer-based models such as Autoformer \cite{wu2021autoformer} and FEDformer \cite{zhou2023fedformer} have demonstrated superior performance in complex temporal modeling. (5) Ensemble methods, which combine multiple forecasting models, have consistently shown robust accuracy improvements \cite{januschowski2020criteria}. (6) Additionally, large language models (LLMs) have been explored for time-series forecasting, leveraging transfer learning from pre-trained architectures \cite{chen2023gpt}.

\textbf{Limitations of Autoregressive Models in Forecasting:}
One of the key limitations in many forecasting applications is the reliance on autoregressive components, which can pose significant challenges. Wang \textit{et al.} \cite{wang2019review}  found that 87\% of DL-based short-term load forecasting models incorporate autoregressive elements. Similarly, Legault \textit{et al.} \cite{legault2022deep} reviewed 142 electricity market forecasting studies and found that 93\% relied on lagged target variables. While effective in stable environments, these models suffer when applied to datasets with structural breaks or regime shifts, as documented by the Monash Time Series Forecasting Archive \cite{godahewa2021monash}. This issue is particularly critical in electricity systems undergoing rapid transformations, where models must generalize well to unseen conditions without relying solely on historical patterns.

0Despite skepticism regarding ML models in long-horizon forecasting, recent research has proposed novel methodologies that demonstrate improved performance over extended periods \cite{lim2021temporal, rangapuram2018deep, li2022efficiently, montero2020fforma}. Advances in theoretical time-series modeling have contributed to better forecasting techniques, as outlined in \cite{du2023tsfpaper}. However, the success of any forecasting approach depends on its implementation, dataset characteristics, and problem constraints. Our study explores these challenges by systematically evaluating forecasting models under constrained settings. Given the dataset's limitations—including a lack of autoregressive target values and a minimal set of exogenous variables—we analyze both statistical and ML-based models to ensure a fair, transparent, and data-driven approach.

\label{relatedwork}

\section{Dataset Analysis}
\begin{table*}[htbp]
  \centering
  \caption{Descriptive Statistics of the Dataset}
  \label{tab:statistics1}
  \resizebox{\textwidth}{!}{%
  \begin{tabular}{ccccccccccccc}
    \toprule
    & \multirow{2}{*}{Time} & Load & \multicolumn{5}{c}{Temperature (\textdegree C)} & \multicolumn{5}{c}{Global Horizontal Irradiance (W/m\textsuperscript{2})} \\
    \cmidrule(lr){3-3} \cmidrule(lr){4-8} \cmidrule(lr){9-13}
    & & & Site 1 & Site 2 & Site 3 & Site 4 & Site 5 & Site 1 & Site 2 & Site 3 & Site 4 & Site 5 \\
    \midrule
    $\mu$ 
    & Year 1 & 2162.82 & 17.31 & 17.17 & 18.27 & 17.72 & 17.60 & 225.80 & 222.47 & 226.91 & 227.06 & 228.39 \\
    & Year 2 & 2145.42 & 16.72 & 16.47 & 17.80 & 17.06 & 16.84 & 221.24 & 218.81 & 225.35 & 223.31 & 225.08 \\
    \midrule
    $\sigma$ 
    & Year 1 & 465.77 & 4.86 & 4.51 & 7.36 & 4.75 & 5.84 & 305.18 & 301.51 & 307.34 & 306.63 & 308.05 \\
    & Year 2 & 406.48 & 4.48 & 4.18 & 7.02 & 4.35 & 5.37 & 301.69 & 298.73 & 306.39 & 304.56 & 306.06 \\
    \midrule
    Median 
    & Year 1 & 2072.00 & 17.50 & 17.40 & 17.30 & 17.80 & 17.30 & 12.00 & 12.00 & 12.00 & 12.00 & 13.00 \\
    & Year 2 & 2096.00 & 16.60 & 16.40 & 17.20 & 17.00 & 16.60 & 12.00 & 11.00 & 11.00 & 11.00 & 11.00 \\
    \midrule
    Min 
    & Year 1 & 1101.00 & 1.90 & 2.90 & -0.50 & 2.60 & 0.90 & 0.00 & 0.00 & 0.00 & 0.00 & 0.00 \\
    & Year 2 & 1027.00 & 4.60 & 4.50 & 0.20 & 3.90 & 2.90 & 0.00 & 0.00 & 0.00 & 0.00 & 0.00 \\
    \midrule
    Max 
    & Year 1 & 4397.00 & 36.60 & 31.90 & 43.00 & 36.60 & 39.70 & 1037.00 & 1028.00 & 1041.00 & 1047.00 & 1049.00 \\
    & Year 2 & 3808.00 & 30.10 & 30.20 & 38.80 & 31.20 & 33.60 & 1031.00 & 1024.00 & 1035.00 & 1042.00 & 1045.00 \\
    \midrule
    $\gamma_1$ 
    & Year 1 & 1.18 & 0.09 & -0.00 & 0.45 & 0.20 & 0.42 & 1.07 & 1.09 & 1.09 & 1.08 & 1.08 \\
    & Year 2 & 0.67 & 0.06 & -0.02 & 0.37 & 0.11 & 0.27 & 1.11 & 1.11 & 1.09 & 1.11 & 1.10 \\
    \midrule
    $\gamma_2$ 
    & Year 1 & 2.10 & -0.14 & -0.33 & -0.26 & -0.02 & 0.02 & -0.25 & -0.19 & -0.20 & -0.23 & -0.23 \\
    & Year 2 & 0.83 & -0.36 & -0.45 & -0.45 & -0.30 & -0.34 & -0.17 & -0.13 & -0.21 & -0.17 & -0.19 \\
    \bottomrule
  \end{tabular}%
  }
\end{table*}

The dataset was provided by the ESD Competition organizers as the PG\&E utility load dataset. It includes two years of training data and one year of test data. The dataset contains temporal features (e.g., year, month), electricity load, and exogenous variables such as temperature and global horizontal irradiance (GHI) from five different sites. However, the exact years of the dataset and the specific locations of the sites are not disclosed. Additionally, the use of any external exogenous variables beyond those provided in the dataset is not permitted. Moreover, a key constraint of the competition is that all hourly exogenous variables are only available on a 'day-ahead' basis, meaning they can be used for forecasting the following day. Specifically, day D's hourly load prediction can utilize any or all the exogenous variables of day D, but not of day D+1 or any future day's exogenous information. This restriction prevents the use of next-day exogenous variables to forecast the current day's load.

\subsection{Statistical Characteristics of the Dataset}
The descriptive statistics of the dataset are provided in Table \ref{tab:statistics1}. We note that the statistics for Load for Year 1 imply a rather 'scattered' dataset. In particular, what is noticeable is that the standard deviation ($\sigma$), as well as third ($\gamma_1$) and fourth ($\gamma_2$) statistical moments, are larger for Load for Year 1 than for the same quantity of Year 2. The maximum data point is also larger by more than 10\%. We can observe a somewhat similar pattern in temperature and GHI data, though those differences between years are much less prominent than for the Load. This can already lead us to a few important conclusions right away. (1) Firstly, the Load fluctuations are somewhat different for two years. This means while we can create test cases of predicting one year using the training set as another, it should be kept in mind that two years have varying patterns. Training using data from both years is crucial for creating final predictions as they capture the varying relationships across years. (2) Secondly, the exogenous variables variability shows that there might be other factors impacting time-series predictions. The Load varies across years, while temperature and GHI are to a much smaller extent suggesting that the difference is most likely caused by some other external factors or a high sensitivity to the exogenous variables. 

\begin{figure}
    \centering
    \includegraphics[width=0.9\linewidth]{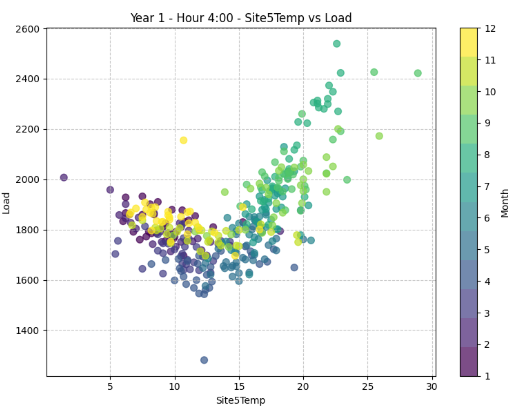}
    \caption{Load and Site 5 Temperature for Year 1.}
    \label{fig:fig1}
\end{figure}

\begin{figure}
    \centering
    \includegraphics[width=0.9\linewidth]{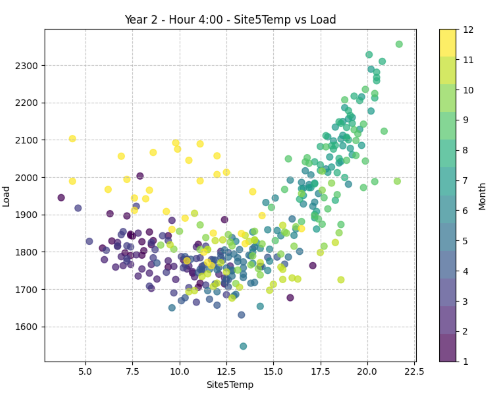}
    \caption{Load and Site 5 Temperature for Year 2.}
    \label{fig:fig2}
\end{figure}

\subsection{Exploring Relationships Between Load and Temperature}
To better understand the dependencies between electricity load and temperature, we conduct an hourly analysis inspired by \cite{hu2023data}. Figure \ref{fig:fig1}, illustrates these relationships, showing clear seasonal effects through two-segment linear relationships: 
\begin{itemize}
    \item \textbf{Winter months} (left heating tail): During colder months, load increases as temperature decreases due to heating demand.
    \item \textbf{Summer months} (right cooling tail): In contrast, during hotter months, load rises with temperature due to cooling needs.
\end{itemize}
The region in between can be called a 'comfort zone' and is prominent in spring months. The particular graph for Hour 4 demonstrates a strong linear relationship, suggesting a regular segmented (2 segments for summer and winter) regression fit. Nevertheless, if we look at the same figure for Year 2 (Figure \ref{fig:fig2}), we observe that these relationships are not entirely consistent. For example:  Firstly, the December 'heating tail' (yellow scatter points) does not coincide with other winter months as it did for Year 1. Moreover, the 'comfort zone' temperature range might be not necessarily the same for both years. This demonstrates the differences between Load behavior in the training data for the years provided. 

For this particular hour, the relationships of the two segments look roughly linear, however, the same figure looks much more scattered for other instances; especially for the instances during the sunlight hours, in which solar power starts playing a crucial role in defining the load dynamics. Further on, picking the 'comfort zone' temperature range can be also challenging as this range does not demonstrate any linear relationships and can belong both to the 'heating tail' and the 'cooling tail' depending on the choice of the breaking point. This suggests that models other than segmented linear regression can demonstrate much better performance. The model can potentially capture the 'comfort zone' relationships better as well as the regime-switching behavior between years. Hence, we iterate through various models and draw a rigorous comparison.








\label{dataset}

\section{Proposed Methodology}

As we take inspiration from \cite{hu2023data}, we create hourly models and adopt them as our overall approach. We also handle the range of the exogenous variables by taking their PCA. We compare various models, ranging from simple statistical techniques to SOA complex architectures. 

\subsection{PCA Feature Engineering}
To address the challenge of handling exogenous variables from multiple sites, we employ Principal Component Analysis (PCA) to reduce dimensionality while retaining the majority of variance in the data. Instead of including all temperature and GHI variables, which may introduce redundancy, multicollinearity, and computational complexity, we extract principal components that capture the most significant variance across the dataset.

Since no single site consistently exhibits the highest correlation with load, relying on one representative site is not ideal. We, thus decide by two primary approaches on how to handle feature selection:-

\begin{enumerate}

    \item Principal Component Analysis (PCA): This method extracts the most significant components from temperature and GHI separately, ensuring that a few principal components capture the majority of the variance. This allows us to replace multiple correlated features with a lower-dimensional representation that maintains relevant information.
    \item Variance Inflation Factor (VIF)-Based Feature Selection: This approach evaluates multicollinearity by computing the Variance Inflation Factor (VIF) for each feature. 
    VIF quantifies how much a feature is correlated with other independent variables in a dataset. It is calculated as: 
    \begin{align*}
        VIF_{i} = \frac{1}{1-R_i^2}
    \end{align*}
    where, $R_i$ is the coefficient of determination (R-squared) when feature $i$ is regressed on all other features.

    A higher VIF means the feature is highly correlated with other independent variables, leading to instability in regression models. Features with excessively high VIF values (typically >10) contribute significantly to redundancy and can be removed.
\end{enumerate}

\begin{table}[]
    \centering
    \caption{Variance Inflation Factor (VIF) results for temperature and GHI features before PCA.}
    \label{tab:VIFbeforePCA}
    \begin{multicols}{2}
        \centering
        \begin{tabular}{cc}
        \toprule
        \textbf{Feature} & \textbf{VIF} \\ \midrule
        Site-1 Temp  & 449.441269  \\
        Site-2 Temp  & 458.227488  \\
        Site-3 Temp  & 114.246325  \\
        Site-4 Temp  & 704.495510  \\
        Site-5 Temp  & 463.611923  \\
        Load         & 20.230474   \\ \bottomrule
        \end{tabular}
        \columnbreak
        \hspace{2.5pt}  
        \vrule width 1pt  
        \hspace{2.5pt}  
        \centering
        \begin{tabular}{cc}
        \toprule
        \textbf{Feature} & \textbf{VIF} \\ \midrule
        Site-1 GHI  & 279.130327  \\
        Site-2 GHI  & 189.029012  \\
        Site-3 GHI  & 144.154615  \\
        Site-4 GHI  & 1067.949872  \\
        Site-5 GHI  & 1630.840315  \\
        Load        & 1.489604   \\ \bottomrule
        \end{tabular}
    \end{multicols}
\end{table}
Before applying PCA, we computed the VIF for the temperature and GHI variables across all sites. The results as seen in Table \ref{tab:VIFbeforePCA} demonstrate severe multicollinearity. Since VIF values exceeding 10 indicate high collinearity, including all site-specific temperature and GHI features would introduce redundancy. Applying PCA reduces this issue significantly.

\begin{figure}[H]
    \centering
    \includegraphics[width=0.6\linewidth]{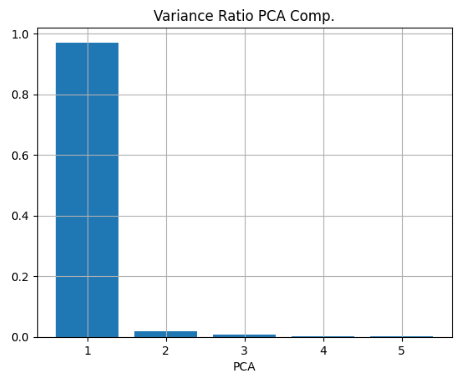}
    \caption{PCA components.}
    \label{fig:PCA}
\end{figure}

From Figure \ref{fig:PCA}, on applying PCA, we observe that the first principal component itself explains 99\% of the variance in the dataset. This suggests that a single principal component is sufficient to retain the majority of the information from the temperature and GHI variables, further validating our dimensionality reduction approach. After PCA transformation, we recompute the VIF values and witness that they drop significantly, confirming the effectiveness of PCA in reducing multicollinearity while retaining essential information (as seen in Table \ref{tab:vif_pca}).

\begin{table}[htbp]
    \centering
    \caption{VIF results after PCA.}
    \label{tab:vif_pca}
    \begin{tabular}{l c}
        \toprule
        \textbf{Feature} & \textbf{VIF} \\
        \midrule
        Load      & 1.013176 \\
        PCA\_Temp & 1.683112 \\
        PCA\_GHI  & 1.672382 \\
        \bottomrule
    \end{tabular}
\end{table}

\subsection{Our Model}
\label{sec:ourModel}
After PCA, our forecasting approach decomposes the 24-hour load prediction task into 24 independent hourly models, each specialized in predicting the load for a specific hour of the day. This ensures that each model captures the distinct temporal dependencies and variability associated with its respective hour. Let- \\
a) $y_{t,h}$ denote the load at time $t$ for hour $h$, where $h \in \{0,1,2,\dots,23\}$. \\
b) $\mathbf{X}_{t,h}$ represents the exogenous features, including PCA-transformed temperature and GHI, as well as temporal variables.\\
c) $f_h(\cdot)$ is the function representing the predictive model trained for each specific hour $h$.\\

\noindent Each model $f_h $ is trained independently on historical data:
\begin{equation}
    y_{t,h} = f_h(\mathbf{X}_{t,h}) + \epsilon_{t,h}
\end{equation}
\noindent where $\epsilon_{t,h}$ represents the residual error for the respective hour $h$.\\

\noindent \textbf{Stacking Predictions:}
Once all 24 hourly models are trained, they are used to generate hourly predictions for a full day. The final daily prediction sequence is constructed as:

\begin{equation}
    \hat{Y}_t = \{ \hat{y}_{t,0}, \hat{y}_{t,1}, \dots, \hat{y}_{t,23} \}
\end{equation}

where:

\begin{equation}
    \hat{y}_{t,h} = f_h(\mathbf{X}_{t,h}), \quad h \in \{0,1,2,\dots,23\}
\end{equation}

This process is iteratively repeated for each day in the forecasting horizon, ultimately constructing a full-year prediction.\\

\noindent \textbf{Advantages of Hourly Model Decomposition:}
\begin{itemize}
    \item \textbf{Capturing Hour-Specific Patterns:} Each model learns the distinct behavior of load at a given hour, leading to improved forecast accuracy.
    \item \textbf{Reduced Model Complexity:} Instead of training a single large model, the task is decomposed into multiple smaller, specialized regressions.
    \item \textbf{Scalability:} Since each hourly model is independent, training can be efficiently parallelized. (Note that, later we also introduce autoregressive factors for improvements, which allow dependencies between models).
\end{itemize}

By leveraging this methodology, our forecasting models capture hour-specific dependencies while maintaining computational efficiency and interoperability. 

The models ($f_h$) we are comparing are the following: 1) Piecewise Linear Regression; 2) Polynomial Regression; 3) XGBoost; 4) Random Forest (RF); 5) Multi-layer Perceptron Neural Networks (MLP NNs); 6) Long Short-Term Memory (LSTM); 7) Gaussian Process (GP); 8) Transformer; 9) Neural Hierarchical Interpolation for Time Series Forecasting (NHITS); 10) Temporal Convolutional Networks (TCN); 11) Temporal Fusion Transformer (TFT); 12) TimeGPT (pre-trained LLM model).

\begin{figure}
    \centering
    \includegraphics[width=\linewidth]{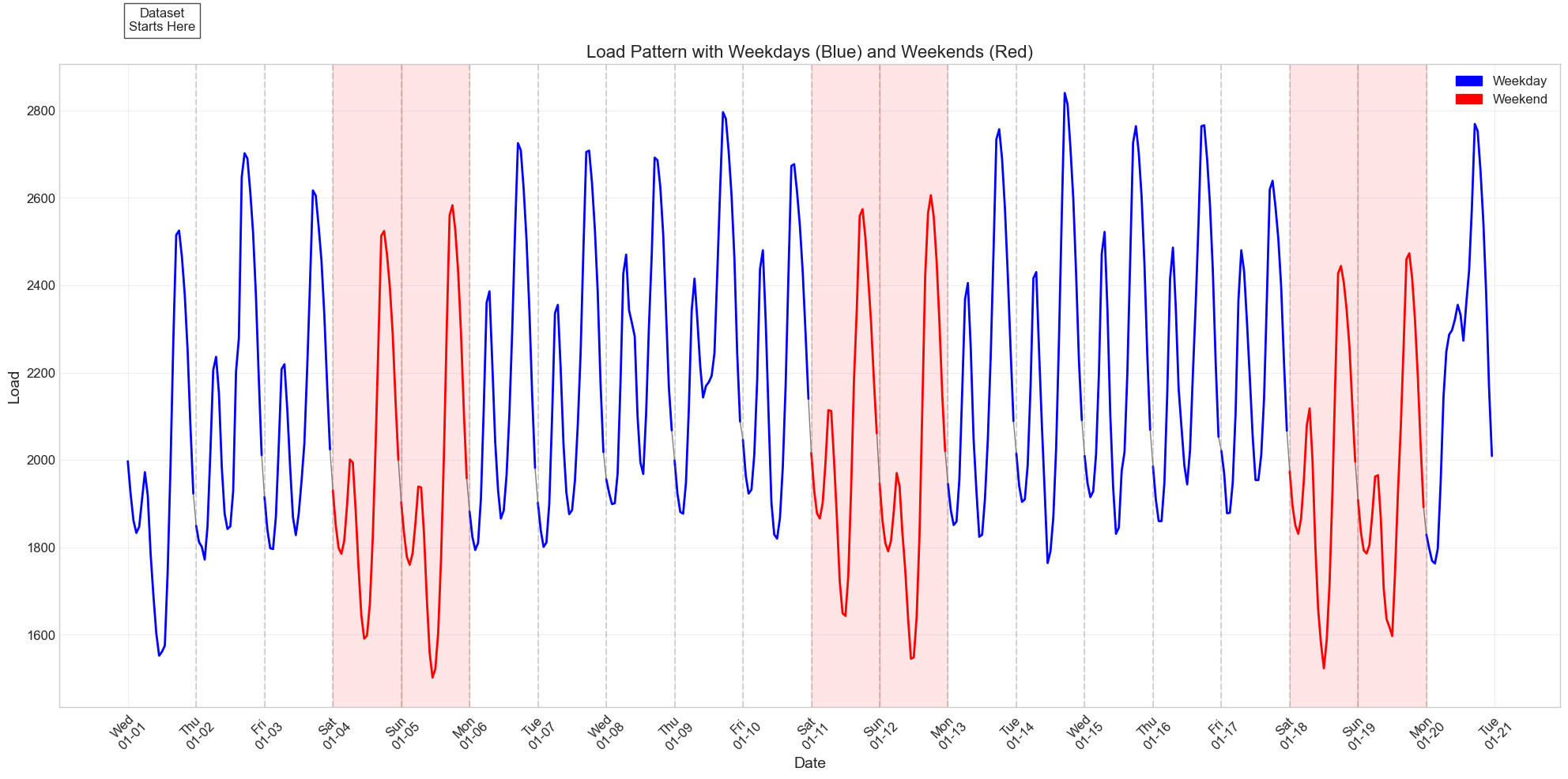}
    \caption{Weekdays vs Weekends from Load data}
    \label{fig:weekends}
\end{figure}

Additionally, the exogenous variables ($\mathbf{X}_{t,h}$) included for the initial model comparison are as follows: 1) PCA temperature; 2) PCA GHI; 3) Monthly dummy variables; 4) Holiday dummy variables (represents 1 on typical days off such as President's Day and 0 otherwise); 5) Weekend dummy variable \footnote{Although the year is not explicitly disclosed in the dataset, our assumption is based on the following reasoning: The first year is a leap year and is known to start on January 1. Given this information, we infer that the year starts on a Wednesday. According to historical records \url{https://en.wikipedia.org/wiki/Leap_year_starting_on_Wednesday}, in the 21st century, the only leap year starting on a Wednesday is 2020. Therefore, we deduce that the three years in the dataset correspond to 2020, 2021, and 2022. This assumption holds for all the temporal variables used in our analysis, including weekend identification, monthly dummies, and holiday indicators.} (this is not using additional exogenous variables but rather leveraging temporal features - the above can be inferred by directly looking at Figure \ref{fig:weekends} i.e. a plot of load vs date - where we find dips during the weekends marked in red).

\subsection{Model Evaluation}
To compare different models we create various test cases. The test cases are meant to leverage different aspects of the training set without looking into the testing set. The three tests we create are:
\begin{enumerate}
    \item Using Year 1 for training and predicting Year 2;
    \item using Year 2 for training and predicting Year 1;
    \item cross-validation (CV) across the entire data set with 5 fold and extended window. 
\end{enumerate}

The test cases are computed as hourly models as described in the previous section \ref{sec:ourModel}, and the predictions and error metrics are computed sequentially.Each of the 24 models, spanning all architectures and test cases, is optimized using Optuna's hyperparameter tuning framework, which efficiently searches for the best parameters through a structured grid search approach.

Once the most accurate model is identified, we also proceed to consider lagged and leading features of exogenous variables for further improvements. 

The key evaluation metrics used for deciding the final model are provided below. The following notation is used: $y_i$ is the actual value, $\hat{y}_i$ is the predicted value, $\bar{y}$ is the mean of the actual values, $n$ is the number of observations.

\begin{equation}
R^2 = 1 - \frac{\sum_{i=1}^{n} (y_i - \hat{y}_i)^2}{\sum_{i=1}^{n} (y_i - \bar{y})^2}
\end{equation}

\begin{equation}
\text{RMSE} = \sqrt{\frac{1}{n} \sum_{i=1}^{n} (y_i - \hat{y}_i)^2}
\end{equation}

\begin{equation}
\text{MAPE} = \frac{100\%}{n} \sum_{i=1}^{n} \left| \frac{y_i - \hat{y}_i}{y_i} \right|
\end{equation}

\begin{equation}
\text{sMAPE} = \frac{100\%}{n} \sum_{i=1}^{n} \frac{|y_i - \hat{y}_i|}{(|y_i| + |\hat{y}_i|)/2}
\end{equation}

\label{methodology}

\section{Experiments and Results}

\begin{table*}[htbp]
  \centering
  \caption{Model Comparison Results}
  \label{tab:modelcomp}
  \resizebox{\textwidth}{!}{%
    \begin{tabular}{llcccccccccccc}
      \toprule
      & & \multicolumn{12}{c}{Models} \\
      \cmidrule(lr){3-14}
      Metric & Test Case & Piecewise & Poly & XGBoost & RF & MLP & GP & LSTM & TF & NHITS & TCN & TFT & TimeGPT \\
      \midrule
      \multirow{3}{*}{$R^2$}
      & Year 1 $\rightarrow$ Year 2 & -3.2 & 0.6 & \textbf{0.8} & 0.8 & 0.8 & 0.8 & 0.7 & 0.7 & 0.7 & 0.7 & 0.3 & 0.18 \\
      & Year 2 $\rightarrow$ Year 1 & 0.7 & 0.8 & \textbf{0.9} & 0.8 & 0.9 & 0.9 & 0.7 & 0.7 & 0.6 & 0.7 & 0.2 & 0.32 \\
      & Both years & -7054e3 & -750e6 & \textbf{0.7} & 0.6 & 0.5 & 0.5 & 0.5 & 0.5 & 0.3 & 0.4 & 0.4 & $\times$ \\
      \midrule
      \multirow{3}{*}{RMSE}
      & Year 1 $\rightarrow$ Year 2 & 833.2 & 263.4 & \textbf{159.6} & 174.8 & 160.6 & 168.9 & 220.9 & 239.5 & 233.6 & 222.8 & 329.3 & 369.03 \\
      & Year 2 $\rightarrow$ Year 1 & 242.0 & 181.6 & 178.6 & 189.6 & \textbf{170.7} & 177.2 & 254.8 & 270.7 & 294.4 & 237.7 & 408.0 & 383.18 \\
      & Both years & 1e6 & 12e6 & \textbf{263.7} & 286.5 & 312.5 & 308.1 & 311.6 & 309.9 & 368.5 & 334.9 & 348.7 & $\times$ \\
      \midrule
      \multirow{3}{*}{MAPE}
      & Year 1 $\rightarrow$ Year 2 & 6.8 & 6.9 & \textbf{5.5} & 5.9 & 5.6 & 5.9 & 7.5 & 8.3 & 8.3 & 7.6 & 11.5 & 14.31 \\
      & Year 2 $\rightarrow$ Year 1 & 6.0 & 5.7 & \textbf{5.6} & 5.9 & 5.8 & 5.9 & 8.5 & 8.9 & 9.3 & 8.0 & 13.1 & 14.79 \\
      & Both years & 4819.8 & 37e3 & \textbf{7.4} & 8.0 & 9.0 & 8.1 & 9.3 & 9.2 & 10.5 & 9.7 & 10.1 & $\times$ \\
      \midrule
      \multirow{3}{*}{sMAPE}
      & Year 1 $\rightarrow$ Year 2 & 5.8 & 7.0 & \textbf{5.5} & 5.9 & 5.6 & 6.0 & 7.5 & 8.2 & 8.2 & 7.6 & 10.9 & 13.19 \\
      & Year 2 $\rightarrow$ Year 1 & 5.9 & 5.7 & \textbf{5.6} & 5.8 & 5.7 & 5.8 & 8.4 & 8.8 & 9.3 & 7.9 & 12.8 & 13.63 \\
      & Both years & 13.0 & 14.0 & \textbf{7.7} & 8.3 & 9.3 & 8.6 & 9.6 & 9.6 & 10.9 & 10.2 & 10.5 & $\times$ \\
      \bottomrule
    \end{tabular}%
  }
\end{table*}

\begin{figure*}
    \centering
    \includegraphics[width=0.8\linewidth]{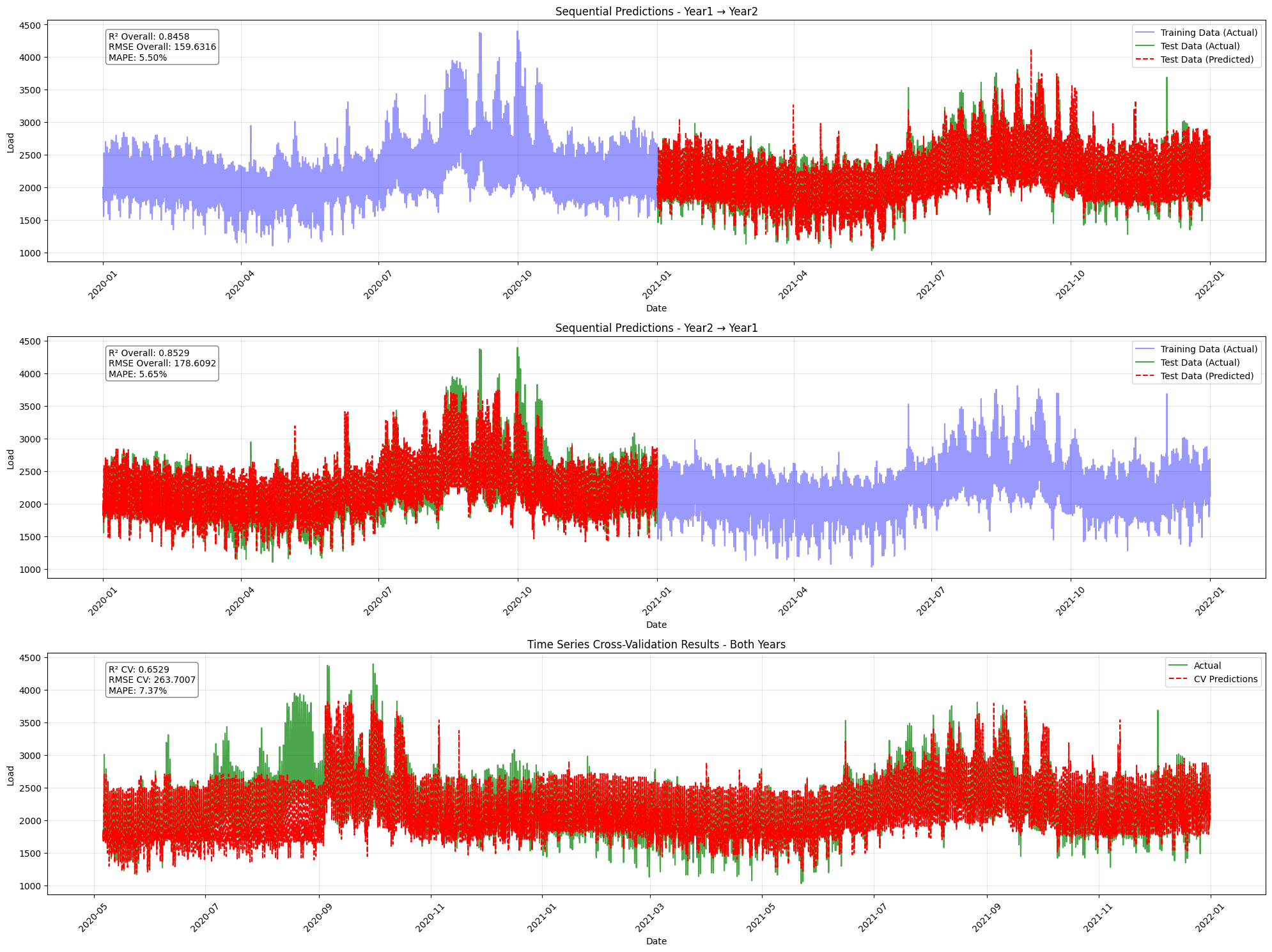}
    \caption{Test Cases for XGBoost Model}
    \label{fig:xgboost}
\end{figure*}



\subsection{Model Comparisons}
We present the model comparison results in Table \ref{tab:modelcomp} and highlight several key observations :

\begin{enumerate}
    \item While regression models can demonstrate relatively strong performance, they tend to collapse during cross-validation. This behavior is expected, as regression models trained on limited data often struggle with outliers and make poor predictions for extreme cases. Consequently, the MAPE error spikes significantly, and the $R^2$ metric remains low, indicating poor generalization.
    \item Complex model architectures do not provide substantial performance improvements across any test cases. This is understandable, as these models are constrained by a limited training dataset and a restricted set of exogenous features. Despite this, we observe a positive outcome, as nearly all models achieve accuracy within a 10-15\% error margin, suggesting that our overall forecasting approach is robust. (Check \textsc{Appendix} \ref{timegptt} for a discussion on the TimeGPT model, as it is modeled differently from our approach).
    From a computational perspective, complex machine learning architectures demand significantly higher training time, ranging from one hour to several days (e.g., TFT models and TimeGPT), whereas most other models complete training within an hour. Given that these architectures do not yield proportionate performance gains, the computational cost does not justify their use.
\end{enumerate}

\begin{table*}[htbp]
  \centering
  \caption{XGBoost Improvement Results}
  \label{tab:improvement}
  \resizebox{\textwidth}{!}{%
    \begin{tabular}{llccccccccccc}
      \toprule
      & & \multicolumn{11}{c}{Models (All XGBoost)} \\
      \cmidrule(lr){3-13}
      Metric & Test Case & Baseline & Lag1 & Lag2 & Lead1 & Lag5 & \makecell{Lag1\\+Lead1} & Lag3 & \makecell{Lag2\\+Lead2} & Lead2 & \makecell{Lag3\\+Lead2} & \makecell{Lag5\\+Lead2} \\
      \midrule
      \multirow{3}{*}{$R^2$}
      & Year 1 $\rightarrow$ Year 2 & 0.84 & 0.84 & 0.85 & 0.84 & 0.85 & 0.84 & 0.84 & 0.85 & 0.84 & 0.85 & \textbf{0.85} \\
      & Year 2 $\rightarrow$ Year 1 & 0.85 & 0.85 & 0.86 & 0.86 & 0.86 & 0.86 & 0.86 & 0.86 & 0.86 & 0.86 & \textbf{0.87} \\
      & Both Years & 0.65 & 0.67 & 0.67 & 0.66 & 0.67 & 0.68 & 0.67 & 0.70 & 0.67 & 0.67 & \textbf{0.68} \\
      \midrule
      \multirow{3}{*}{RMSE}
      & Year 1 $\rightarrow$ Year 2 & \textbf{161.07} & 162.93 & 163.64 & 163.88 & 166.35 & 166.45 & 166.56 & 166.70 & 167.00 & 168.28 & 171.27 \\
      & Year 2 $\rightarrow$ Year 1 & 178.93 & 180.30 & 180.74 & \textbf{178.63} & 184.53 & 179.56 & 182.31 & 185.51 & 180.31 & 186.85 & 185.41 \\
      & Both Years & 264.50 & \textbf{258.52} & 264.08 & 265.23 & 269.87 & 262.30 & 264.36 & 261.64 & 265.25 & 275.07 & 276.71 \\
      \midrule
      \multirow{3}{*}{MAPE}
      & Year 1 $\rightarrow$ Year 2 & \textbf{5.54} & 5.55 & 5.54 & 5.59 & 5.59 & 5.64 & 5.61 & 5.67 & 5.67 & 5.69 & 5.73 \\
      & Year 2 $\rightarrow$ Year 1 & \textbf{5.63} & 5.72 & 5.74 & 5.66 & 5.98 & 5.69 & 5.82 & 5.88 & 5.74 & 5.95 & 6.04 \\
      & Both Years & 7.40 & \textbf{7.15} & 7.45 & 7.26 & 7.45 & 7.25 & 7.42 & 7.18 & 7.26 & 7.48 & 7.35 \\
      \midrule
      \multirow{3}{*}{sMAPE}
      & Year 1 $\rightarrow$ Year 2 & \textbf{5.56} & 5.57 & 5.57 & 5.61 & 5.63 & 5.67 & 5.64 & 5.69 & 5.70 & 5.71 & 5.76 \\
      & Year 2 $\rightarrow$ Year 1 & \textbf{5.61} & 5.68 & 5.68 & 5.62 & 5.89 & 5.64 & 5.76 & 5.83 & 5.69 & 5.88 & 5.94 \\
      & Both Years & 7.71 & \textbf{7.44} & 7.73 & 7.57 & 7.77 & 7.53 & 7.70 & 7.47 & 7.55 & 7.79 & 7.72 \\
      \bottomrule
    \end{tabular}%
  }
\end{table*}

Among all models, \textbf{XGBoost emerges as the best-performing model}. Thus, we select it for further iterations. Figure \ref{fig:xgboost} illustrates predictions of the XGBoost model, where the model produces satisfactory predictions, capturing key patterns effectively. However, cross-validation results reveal a different picture—in the first fold, the model fails to capture all dependencies and features, leading to stationary predictions in the initial portion of the test set. As the training window extends, the predictions improve significantly. This phenomenon is not unique to XGBoost, as most models exhibit a major performance drop in the first cross-validation fold, which remains a primary source of error.

In conclusion, XGBoost consistently outperforms other models across all key metrics, demonstrating superior ability to capture relevant patterns and dependencies.

\subsection{Model Improvements - Adding Leading and Lagged Features}

\begin{figure*}
    \centering
    \includegraphics[width=0.7\linewidth]{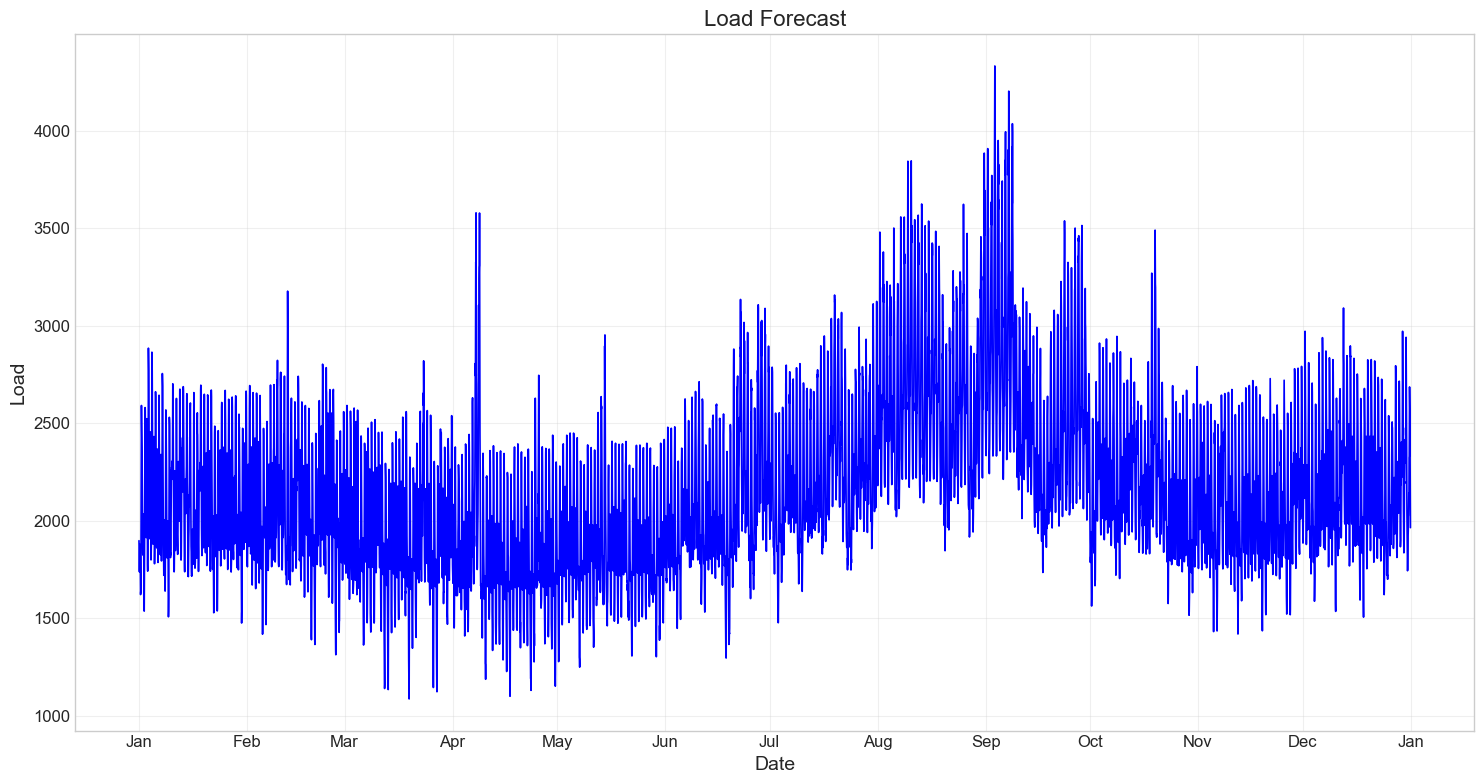}
    \caption{Final Forecast}
    \label{fig:final_forecast}
\end{figure*}

\begin{figure*}
    \centering
    \includegraphics[width=0.7\linewidth]{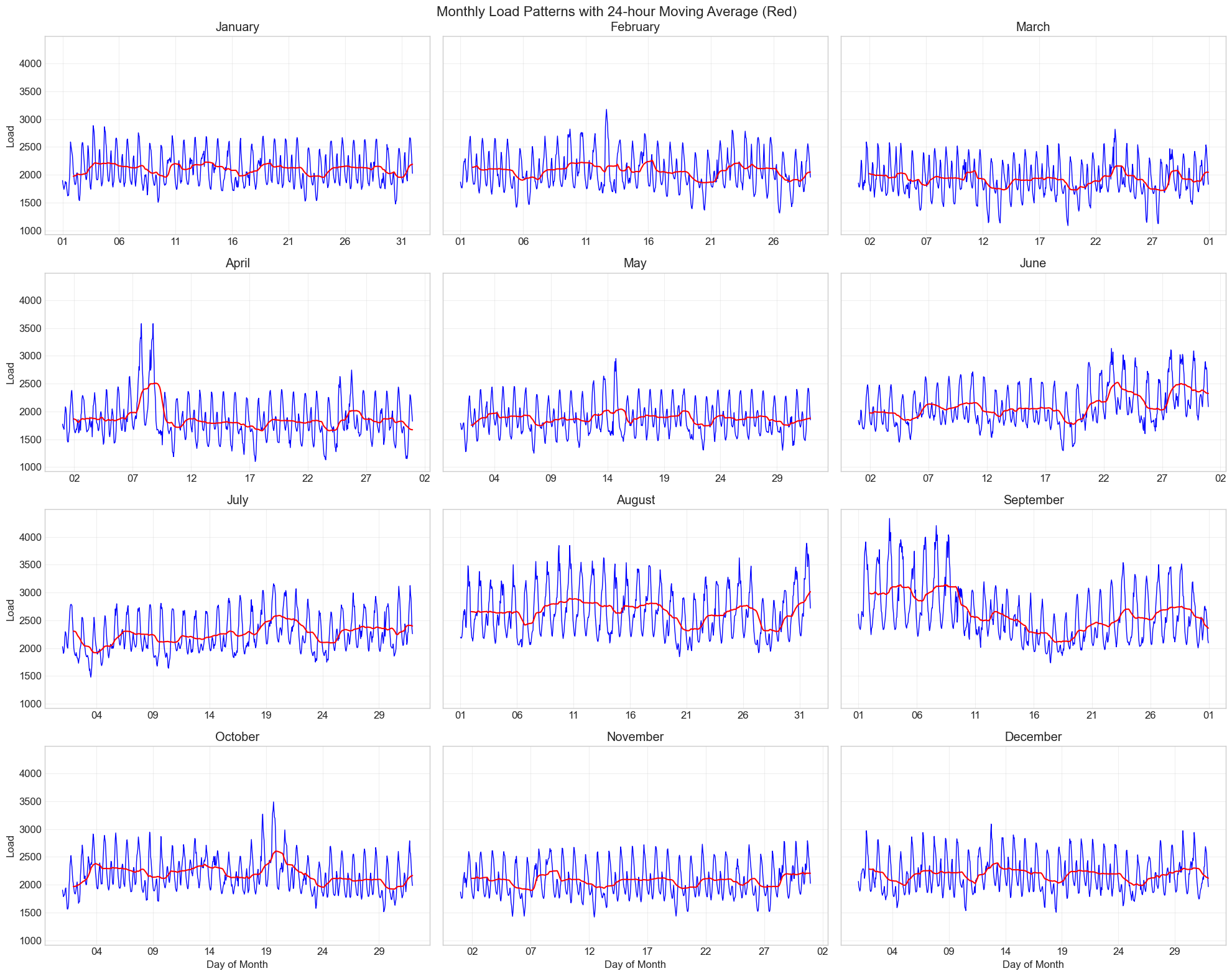}
    \caption{Final Forecast by Month}
    \label{fig:forecast_by_month}
\end{figure*}

We explore various lag and lead combinations to enhance the performance of XGBoost. The only lagging and leading features considered are PCA Temperature and PCA GHI. Given the problem constraints: 
\begin{itemize}
    \item Lagging features are unconstrained for hourly models.
    \item Leading features are restricted to ensure they do not violate daily constraints.
\end{itemize}

Table \ref{tab:improvement} presents the impact of incorporating various lag and lead feature combinations on XGBoost's forecasting performance. The results indicate that performance improvements remain marginal, regardless of the specific lag-lead configuration. Notably, introducing additional exogenous features does not consistently reduce error metrics compared to the Baseline model, which excludes exogenous variable lags.

While some configurations, such as Lag5 + Lead2, achieve the highest $R^2$(0.87 in Year 2 → Year 1), the overall improvement is incremental and does not justify the added computational complexity, especially given its higher RMSE and MAPE values.

Considering both accuracy and efficiency, \textbf{we select Lag1 as our final model for further forecasting iterations}. It exhibits slight but consistent improvements, particularly in cross-validation, while maintaining a balance between performance gains and computational feasibility.\\

We do not conduct an exhaustive analysis of other lagged feature configurations. Potential alternatives could include: Nested lags, Moving averages of exogenous variables, Other feature combinations, etc. However, the impact of these modifications is likely negligible and unlikely to yield significant performance gains. This can be attributed to the fact that the overall temporal temperature dynamics play a minimal role in influencing load shifts. The instantaneous load at any given moment is largely independent of past or future temperatures, as user behavior is driven by real-time decision-making.

In this context, model architecture plays a more critical role in forecast performance than additional lagged features. In other scenarios, \textbf{incorporating lagged values of the actual load could provide substantial improvements}, but in our case, such an approach is not applicable given the problem constraints.

\subsection{Final Model}

Our final model is XGBoost with one lag feature of the PCA of the respective exogenous variables. We optimize each hourly model using a hyperparameter grid search on the entire training set and create sequential predictions. The result is depicted in Figure \ref{fig:final_forecast}. We can see that the load follows generally a satisfactory trend with the peak in the summer and the highest overall load during that season. We can also see the load decline in the spring consistent with the previous observations. We can also observe the load spikes in April, which are most likely associated with the high-temperature spikes (the temperature at some sites goes above 30 degrees Celsius for some of the hours). We can also see consistent dips in the load across the entire dataset, suggesting the reductions in load due to the weekends and potential holidays. We can see the patterns in the load in Figure \ref{fig:forecast_by_month}. The decline in trend is really clear for the month of September, while the gradual increase in July can also be observed. Overall, the load follows satisfactory daily patterns across the predictions. 
\label{experiments}

\section{Conclusion}


This study evaluates the effectiveness of various forecasting models for electricity load prediction, comparing traditional regression-based techniques with advanced deep learning architectures. Our results demonstrate that simpler machine learning models, particularly XGBoost, outperform transformer-based models like TimeGPT in this constrained forecasting setting. Despite their theoretical advantages, deep learning models struggle with limited training data, feature sparsity, and long-term forecasting challenges, leading to error accumulation and performance degradation over extended horizons.

Through systematic experimentation, we show that model selection plays a far more critical role than extensive feature engineering. While lagging and leading exogenous variables offer only marginal improvements, our final model—XGBoost with a single lagged feature (Lag1)—achieves the best trade-off between accuracy and computational efficiency. This highlights that in scenarios with limited exogenous inputs, interpretable and computationally efficient models remain preferable over deep learning alternatives.

Our findings underscore the ongoing challenges in long-term electricity load forecasting and raise important questions about the applicability of deep learning for such tasks. Additionally, large-scale pre-trained models, such as LLM-based forecasting architectures, offer a potential avenue for future exploration—though their effectiveness in constrained settings remains an open question.

Ultimately, our study contributes to the broader discussion on forecasting model selection, reinforcing that model complexity does not always translate to better performance. Practical forecasting solutions must be tailored to dataset characteristics, ensuring that simplicity, interpretability, and efficiency remain key considerations in model deployment.
\label{conclusion}

\bibliographystyle{plain}
\begin{figure*}
    \centering
    \includegraphics[width=\linewidth]{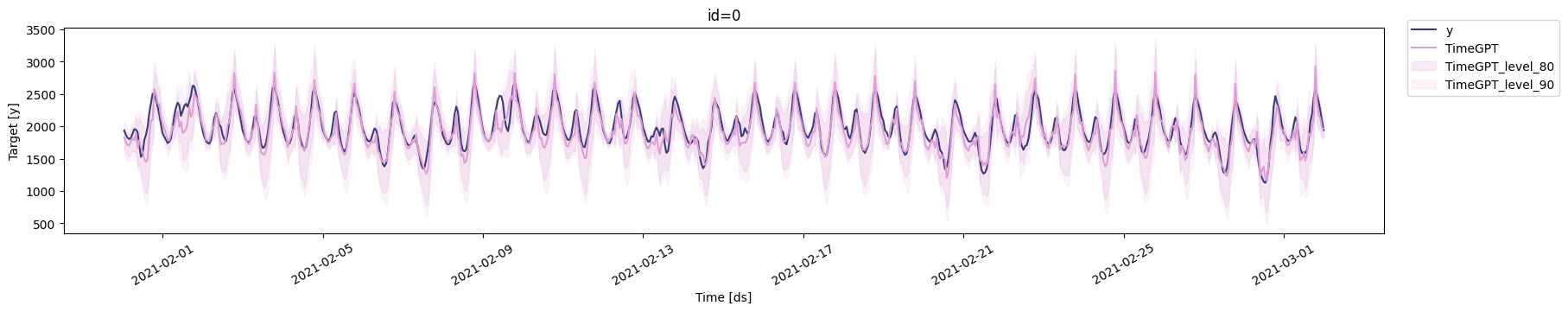}
    \label{fig:enter-label}
\end{figure*}
\begin{figure*}
    \centering
    \includegraphics[width=\linewidth]{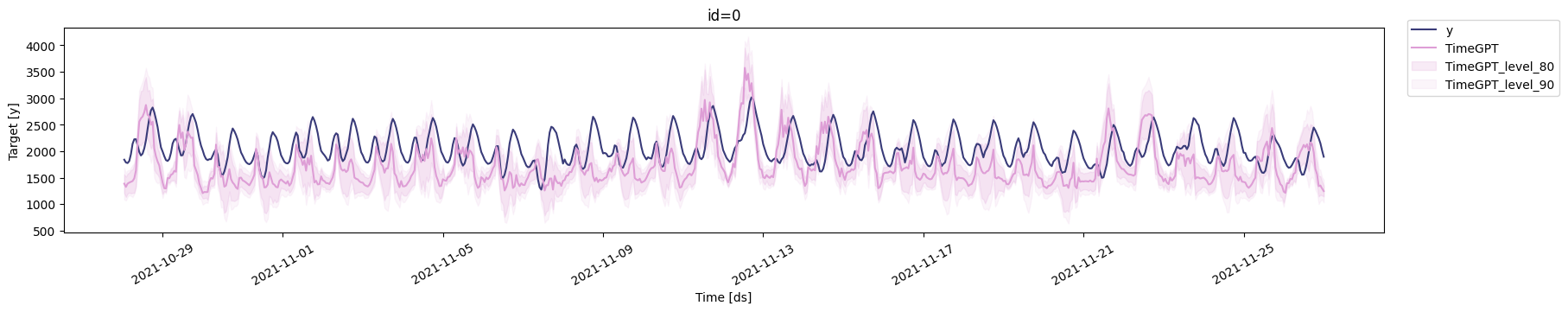}
    \caption{Long-Horizon TimeGPT-1 Forecasting}
    \label{fig:timegpt}
\end{figure*}


\begin{thebibliography}{10}

\bibitem{arias2022probabilistic}
Mariana Arias~Chao and Volodymyr Kuleshov.
\newblock Probabilistic time series forecasting with recurrent neural networks for intermittent demand.
\newblock {\em IEEE Transactions on Neural Networks and Learning Systems}, 33(6):2620--2631, 2022.

\bibitem{benidis2020neural}
Konstantinos Benidis, Syama~Sundar Rangapuram, Valentin Flunkert, Bernie Wang, Danielle Maddix, Caner Turkmen, Jan Gasthaus, Michael Bohlke-Schneider, David Salinas, Lorenzo Stella, et~al.
\newblock Neural forecasting: Introduction and literature overview.
\newblock {\em arXiv preprint arXiv:2004.10240}, 2020.

\bibitem{chen2023gpt}
Wei Chen, Jingyu Chen, and Jianmin Zhang.
\newblock Gpt4ts: Prompt learning for time series forecasting.
\newblock {\em arXiv preprint arXiv:2310.13567}, 2023.

\bibitem{du2023tsfpaper}
Dongzhe Du, Wei Chen, Xiuwen Wang, and Jianmin Zhang.
\newblock Tsfpaper: A repository of time series forecasting papers.
\newblock \url{https://github.com/ddz16/TSFpaper}, 2023.
\newblock Accessed: 2025-03-20.

\bibitem{ferc2021staff}
{Federal Energy Regulatory Commission}.
\newblock Staff report on data needs for electricity system planning.
\newblock Staff report, FERC, 2021.

\bibitem{gilliland2019m4}
Michael Gilliland.
\newblock The m4 forecasting competition—a practitioner's view.
\newblock {\em International Journal of Forecasting}, 35(1):161--174, 2019.

\bibitem{godahewa2021monash}
Rakshitha Godahewa, Christoph Bergmeir, Geoffrey~I Webb, Rob~J Hyndman, and Pablo Montero-Manso.
\newblock Monash time series forecasting archive.
\newblock {\em arXiv preprint arXiv:2105.06643}, 2021.

\bibitem{green2015simple}
Kesten~C Green and J~Scott Armstrong.
\newblock Simple versus complex forecasting: The evidence.
\newblock {\em Journal of Business Research}, 68(8):1678--1685, 2015.

\bibitem{haben2016analysis}
Stephen Haben, Colin Singleton, and Peter Grindrod.
\newblock Analysis and clustering of residential customers energy behavioral patterns using robust data mining techniques.
\newblock {\em International Journal of Forecasting}, 32(3):717--730, 2016.

\bibitem{hu2023data}
Yinbo Hu, Michael Waite, Evan Patz, Bainan Xia, Yixing Xu, Daniel Olsen, Naveen Gopan, and Vijay Modi.
\newblock A data-driven approach for the disaggregation of building-sector heating and cooling loads from hourly utility load data.
\newblock {\em Energy Strategy Reviews}, 49:101175, 2023.

\bibitem{hyndman2020large}
Rob~J Hyndman, Earo Wang, and Nikolay Laptev.
\newblock Large-scale unusual time series detection.
\newblock {\em In IEEE International Conference on Data Mining Workshops (ICDMW)}, pages 1616--1619, 2020.

\bibitem{januschowski2020criteria}
Tim Januschowski, Jan Gasthaus, Yuyang Wang, David Salinas, Valentin Flunkert, Michael Bohlke-Schneider, and Laurent Callot.
\newblock Criteria for classifying forecasting methods.
\newblock {\em International Journal of Forecasting}, 36(1):167--177, 2020.

\bibitem{kang2020renewable}
Chongqing Kang, Ning Zhang, Qing Xia, and Yi~Ding.
\newblock {\em Renewable Energy-Storage Systems Integration in Power Grids: Modeling, Control and Optimization}.
\newblock Academic Press, 2020.

\bibitem{keles2016extended}
Dogan Keles, Jonathan Scelle, Florentina Paraschiv, and Wolf Fichtner.
\newblock Extended forecast methods for day-ahead electricity spot prices applying artificial neural networks.
\newblock {\em Applied Energy}, 162:218--230, 2016.

\bibitem{lago2018forecasting}
Jesus Lago, Fjo De~Ridder, and Bart De~Schutter.
\newblock Forecasting spot electricity prices: Deep learning approaches and empirical comparison of traditional algorithms.
\newblock {\em Applied Energy}, 221:386--405, 2018.

\bibitem{lai2018modeling}
Guokun Lai, Wei-Cheng Chang, Yiming Yang, and Hanxiao Liu.
\newblock Modeling long-and short-term temporal patterns with deep neural networks.
\newblock {\em In The 41st International ACM SIGIR Conference on Research \& Development in Information Retrieval}, pages 95--104, 2018.

\bibitem{Lanka}
Vishnu Vardhan~Sai Lanka, Millend Roy, Shikhar Suman, and Shivam Prajapati.
\newblock Renewable energy and demand forecasting in an integrated smart grid.
\newblock In {\em 2021 Innovations in Energy Management and Renewable Resources(52042)}, pages 1--6, 2021.

\bibitem{legault2022deep}
Miguel Legault, Alexandre Ouellet, Gaelle Saint-Hilary, Mathilde Bourdeau, and Barthelemy Ateme-Nguema.
\newblock Deep learning for electricity market forecasting: Current methods, challenges and opportunities.
\newblock {\em Renewable and Sustainable Energy Reviews}, 168:112778, 2022.

\bibitem{li2022efficiently}
Albert~Q Li, Tri Dao, Bryan Lim, Philip Lichtarge, Sercan~O Arik, Michael~Y Li, and Tomas Pfister.
\newblock Efficiently modeling long sequences with structured state spaces.
\newblock {\em arXiv preprint arXiv:2111.00396}, 2022.

\bibitem{lim2021temporal}
Bryan Lim, Sercan~O Arık, Nicolas Loeff, and Tomas Pfister.
\newblock Temporal fusion transformers for interpretable multi-horizon time series forecasting.
\newblock {\em International Journal of Forecasting}, 37(4):1748--1764, 2021.

\bibitem{makridakis2018statistical}
Spyros Makridakis, Evangelos Spiliotis, and Vassilios Assimakopoulos.
\newblock Statistical and machine learning forecasting methods: Concerns and ways forward.
\newblock {\em PloS one}, 13(3):e0194889, 2018.

\bibitem{makridakis2022m5}
Spyros Makridakis, Evangelos Spiliotis, Vassilios Assimakopoulos, Zhi Chen, Anil Gaba, Ilia Tsetlin, and Robert~L Winkler.
\newblock The m5 competition: Background, organization, and implementation.
\newblock {\em International Journal of Forecasting}, 38(4):1325--1336, 2022.

\bibitem{montero2020fforma}
Pablo Montero-Manso, George Athanasopoulos, Rob~J Hyndman, and Thiyanga~S Talagala.
\newblock Fforma: Feature-based forecast model averaging.
\newblock {\em International Journal of Forecasting}, 36(1):86--92, 2020.

\bibitem{nerc2022reliability}
{North American Electric Reliability Corporation}.
\newblock Reliability guideline: Methods for establishing resource adequacy requirements.
\newblock Technical report, NERC, 2022.

\bibitem{nowotarski2013computing}
Jakub Nowotarski and Rafal Weron.
\newblock Computing electricity spot price prediction intervals using quantile regression and forecast averaging.
\newblock {\em Computational Statistics}, 30(3):791--803, 2013.

\bibitem{oreshkin2020n}
Boris~N Oreshkin, Dmitri Carpov, Nicolas Chapados, and Yoshua Bengio.
\newblock N-beats: Neural basis expansion analysis for interpretable time series forecasting.
\newblock {\em arXiv preprint arXiv:1905.10437}, 2020.

\bibitem{petropoulos2022forecasting}
Fotios Petropoulos, Daniele Apiletti, Vassilios Assimakopoulos, Mohamed~Zied Babai, Devon~K Barrow, Souhaib~Ben Taieb, Christoph Bergmeir, Ricardo~J Bessa, Jakub Bijak, John~E Boylan, et~al.
\newblock Forecasting: theory and practice.
\newblock {\em International Journal of Forecasting}, 38(3):705--871, 2022.

\bibitem{rangapuram2018deep}
Syama~Sundar Rangapuram, Matthias~W Seeger, Jan Gasthaus, Lorenzo Stella, Yuyang Wang, and Tim Januschowski.
\newblock Deep state space models for time series forecasting.
\newblock {\em Advances in neural information processing systems}, 31, 2018.

\bibitem{rubaszewski2022probabilistic}
Jakub Rubaszewski, Jakub Nowotarski, and Rafal Weron.
\newblock Probabilistic load forecasting for large-scale distributed energy resources aggregation.
\newblock {\em IEEE Transactions on Smart Grid}, 13(3):2133--2146, 2022.

\bibitem{SAHU2023109025}
Sourav~Kumar Sahu, Millend Roy, Soham Dutta, Debomita Ghosh, and Dusmanta~Kumar Mohanta.
\newblock Machine learning based adaptive fault diagnosis considering hosting capacity amendment in active distribution network.
\newblock {\em Electric Power Systems Research}, 216:109025, 2023.

\bibitem{salinas2020deepar}
David Salinas, Valentin Flunkert, Jan Gasthaus, and Tim Januschowski.
\newblock Deepar: Probabilistic forecasting with autoregressive recurrent networks.
\newblock {\em International Journal of Forecasting}, 36(3):1181--1191, 2020.

\bibitem{sezer2020financial}
Omer~Berat Sezer, Mehmet~Ugur Gudelek, and Ahmet~Murat Ozbayoglu.
\newblock Financial time series forecasting with deep learning: A systematic literature review: 2005--2019.
\newblock {\em Applied Soft Computing}, 90:106181, 2020.

\bibitem{wang2019review}
Yi~Wang, Qixin Chen, Tao Hong, and Chongqing Kang.
\newblock Review of smart meter data analytics: Applications, methodologies, and challenges.
\newblock {\em IEEE Transactions on Smart Grid}, 10(3):3125--3148, 2019.

\bibitem{weron2014electricity}
Rafal Weron.
\newblock Electricity price forecasting: A review of the state-of-the-art with a look into the future.
\newblock {\em International Journal of Forecasting}, 30(4):1030--1081, 2014.

\bibitem{wu2021autoformer}
Haixu Wu, Jiehui Xu, Jianmin Wang, and Mingsheng Long.
\newblock Autoformer: Decomposition transformers with auto-correlation for long-term series forecasting.
\newblock {\em Advances in Neural Information Processing Systems}, 34:22419--22430, 2021.

\bibitem{zhou2023fedformer}
Tian Zhou, Ziqing Ma, Qingsong Wen, Xue Wang, Liang Sun, and Rong Jin.
\newblock Fedformer: Frequency enhanced decomposed transformer for long-term series forecasting.
\newblock {\em In International Conference on Machine Learning}, pages 41269--41286, 2023.

\end{thebibliography}

\appendix
\section{Discussions on TimeGPT}

\textbf{Short-Horizon Forecasting Using TimeGPT-1:}
Due to API constraints and the computational complexity of running 24 independent hourly models, we opted for a sequential daily prediction approach when using TimeGPT-1. Instead of modeling each hour separately, we utilized a daily loop strategy, where predictions for each day were fed into the next day's forecast alongside exogenous variables (PCA Temperature and PCA GHI).

Table \ref{tab:modelcomp} shows the short-horizon forecasting results as it abides by the rules of the competition. The predicted values follow the ground truth well, capturing the overall trend and seasonality. Although minor deviations occur, the model maintains consistency across different test cases. This approach highlights that TimeGPT-1 can capture short-term dependencies reasonably well when applied sequentially, but its effectiveness in long-term forecasting remains questionable.\\

\textbf{Long-Horizon Forecasting Using TimeGPT-1}
To evaluate TimeGPT-1’s performance in long-term forecasting, we conducted an “all-at-once” approach, where the entire year was predicted in one step using PCA-transformed GHI and temperature as exogenous variables.

Figure \ref{fig:timegpt} illustrate the results, highlighting the degradation in forecasting quality from February (the first plot) to November (the bottom plot). This suggests that the model struggles to generalize when required to predict long-term sequences. The visualizations show that predictions become increasingly erratic and lose alignment with the ground truth, as error propagation grows over time.

Key Takeaways from Long-Horizon Forecasting:
\begin{enumerate}
    \item Error accumulation is significant over extended horizons, leading to unreliable long-term predictions.
    \item The lack of autoregressive updates in the model design contributes to forecast degradation.
    \item Deep learning models, including TimeGPT-1, require substantial training data and extensive exogenous inputs to perform well in long-term scenarios.
\end{enumerate}

While TimeGPT-1 offers a promising deep learning-based alternative, its practicality in long-term electricity load forecasting remains limited by computational constraints, feature sparsity, and sensitivity to error propagation. Future research should investigate hybrid approaches that integrate deep learning with structured statistical models to mitigate these challenges.
\label{timegptt}

\end{document}